\documentclass[conference]{IEEEtran}
\IEEEoverridecommandlockouts

\usepackage{graphicx}
\usepackage{amsmath}
\usepackage{booktabs}
\usepackage{hyperref}
\usepackage{multirow}
\usepackage{amssymb}
\usepackage{threeparttable}

\makeatletter
\renewcommand{\footnoterule}{%
  \kern -3pt
  \hrule width \columnwidth height 0.4pt
  \kern 2.6pt
}
\makeatother

\begin{document}

\title{A Micro-Macro Machine Learning Framework for Predicting Childhood Obesity Risk Using NHANES and Environmental Determinants}

\author{
\IEEEauthorblockN{
Eswarasanthosh Kumar Mamillapalli\textsuperscript{*},
and Nishtha Sharma\textsuperscript{*}
}
\IEEEauthorblockA{
eswarasanthoshkumarm@gmail.com,
nishtha\_sharma@nh.gov
}
\thanks{\textsuperscript{*}These authors contributed equally to this work.}
}
\maketitle

\begin{abstract}
Childhood obesity remains a major public health challenge in the United States, strongly influenced by a combination of individual-level, household-level, and environmental-level risk factors. Traditional epidemiological studies typically analyze these levels independently, limiting insights into how structural environmental conditions interact with individual-level characteristics to influence health outcomes.

In this study, we introduce a micro-macro machine learning framework that integrates (1) individual-level anthropometric and socioeconomic data from NHANES and (2) macro-level structural environment features, including food access, air quality, and socioeconomic vulnerability extracted from USDA and EPA datasets. Four machine learning models Logistic Regression, Random Forest, XGBoost, and LightGBM were trained to predict obesity using NHANES microdata. XGBoost achieved the strongest performance.

A composite environmental vulnerability index (EnvScore) was constructed using normalized indicators from USDA and EPA at the state level. Multi-level comparison revealed strong geographic similarity between states with high environmental burden and the nationally predicted micro-level obesity risk distribution. This demonstrates the feasibility of integrating multi-scale datasets to identify environment-driven disparities in obesity risk.

This work contributes a scalable, data-driven, multi-level modeling pipeline suitable for public health informatics, demonstrating strong potential for expansion into causal modeling, intervention planning, and real-time analytics.
\end{abstract}


\section{Introduction}

Childhood obesity has emerged as one of the most pressing public health challenges in the United States, with profound implications for long-term physical, cognitive, and psychosocial development \cite{ogden2014,skinner2018}. Unlike adult obesity, childhood obesity reflects a highly dynamic interplay of early-life exposures, biological growth patterns, household behaviors, and the broader structural environment in which children live \cite{swinburn2011,hill1998,kumanyika2019}. During critical developmental windows, children are uniquely sensitive to environmental stressors such as limited food availability, socioeconomic instability, and poor air quality \cite{leung2019,larson2009,hajat2015,trasande2011,jerrett2010}. As a result, childhood obesity is increasingly viewed not only as a clinical condition, but as a multi-layered societal problem shaped by both micro-level characteristics and macro-level determinants \cite{papas2007,sallis2009,gortmaker2011}.

A substantial body of literature has examined factors associated with pediatric obesity, yet most prior studies tend to focus on either the individual level or the environmental level in isolation \cite{papas2007,larson2009,sallis2009}. Micro-level studies using clinical or survey data typically investigate demographic attributes (e.g., age, race/ethnicity, sex), socioeconomic indicators (e.g., income-to-poverty ratio, parental education), and behavioral risk factors \cite{zhang2021,wang2012,drewnowski2009,wadden2002}. Macro-level studies, in contrast, analyze structural environments, such as food desert prevalence, neighborhood deprivation, pollution exposure, and built-environment quality \cite{powell2007,lake2007,frank2012,hajat2015}. However, very few studies successfully integrate these levels of analysis, largely due to methodological and data-integration barriers \cite{rehkopf2014,kumanyika2019}.

This gap is especially evident in the use of the National Health and Nutrition Examination Survey (NHANES), the gold-standard dataset for individual-level anthropometric and health information \cite{cdc2023,carroll2021}. While NHANES provides precise measurements ideal for modeling childhood obesity, the public-use files intentionally omit state or county identifiers to protect participant anonymity \cite{cdc2023}. This de-identification makes it difficult to directly link child-level health outcomes with county- or state-level environmental metrics. Conversely, large environmental datasets, such as the USDA Food Access Research Atlas and the EPA Air Quality System (AQS), offer rich macro-level indicators of food accessibility, socioeconomic disadvantage, and pollution burden \cite{usda2023,epa2023}, but lack individual health outcomes \cite{larson2009,hajat2015}.

The central challenge, therefore, is to create a computational framework that bridges these heterogeneous data sources operating at different spatial scales, enabling a unified micro-macro understanding of childhood obesity risk. From a machine learning perspective, this requires both robust individual-level predictive modeling and careful aggregation or normalization of environmental indicators to ensure comparability \cite{hastie2009,zou2005}. In addition, because obesity risk is shaped by nonlinear interactions between social conditions and biology, models must be capable of capturing complex structure while still supporting meaningful interpretation of risk drivers \cite{breiman2001,xgboost2016,lightgbm2017,shap2017}.

In this study, we propose a multi-level artificial intelligence framework that integrates:
\begin{enumerate}
    \item \textbf{Micro-level child obesity prediction}, leveraging NHANES anthropometric and socioeconomic variables to model pediatric obesity using multiple machine learning classifiers \cite{cdc2023,carroll2021,breiman2001,xgboost2016,lightgbm2017}.
    \item \textbf{Macro-level environmental vulnerability estimation}, constructed from USDA and EPA datasets to quantify state-level structural burdens relevant to children (food desert exposure, poverty, air quality, and access limitations) \cite{usda2023,epa2023,larson2009,hajat2015}.
    \item \textbf{Cross-scale comparison}, aligning nationally predicted childhood obesity risk with aggregate environmental vulnerability profiles to evaluate structural patterns and identify high-risk states.
\end{enumerate}

This framework contributes to public health informatics by demonstrating how machine learning can fuse multi-granular datasets despite the absence of shared geographic identifiers to generate meaningful insights about childhood obesity \cite{gortmaker2011,kumanyika2019}. More broadly, the work highlights the potential of computational methods to uncover connections between individual health outcomes and the broader social and environmental systems that shape them \cite{swinburn2011,hill1998}. By integrating child-level microdata with state-level macro indicators, this study advances a scalable, data-driven approach for evaluating pediatric health disparities and informing targeted, environment-specific interventions \cite{kumanyika2019,leung2019}.

\section{Related Work}

Research on childhood obesity has established that risk is shaped by interacting biological, behavioral, socioeconomic, and environmental forces rather than a single dominant driver. Large-scale epidemiologic studies have documented persistent U.S. trends and disparities in pediatric obesity prevalence over time, highlighting unequal burden across demographic groups and socioeconomic strata \cite{ogden2014,skinner2018,ford2013}. Longitudinal evidence further shows that excess weight in childhood frequently tracks into adulthood, motivating early identification and prevention strategies \cite{singh2008}. These findings collectively frame childhood obesity as a developmental and structural issue with long-term population-health consequences.

\subsection{Micro-level determinants from clinical and survey data}
A substantial portion of the literature models childhood obesity at the individual (micro) level using survey or clinical cohorts, focusing on demographic and socioeconomic correlates. Socioeconomic status, including income and education, is repeatedly associated with obesity risk, with patterns that reflect broader inequalities in food access, health behaviors, and opportunity structures \cite{zhang2021,wang2012,drewnowski2009,cawley2015}. Behavioral pathways, including dietary quality and fast-food consumption, have also been linked to pediatric weight outcomes \cite{pereira2011,ghosh2014}. While these micro-level analyses provide high-resolution insights into individual risk factors, they often cannot directly quantify structural exposures (e.g., food environment or pollution) that operate upstream of individual choice \cite{papas2007}.

\subsection{Macro-level structural and environmental influences}
Parallel work examines obesity through the lens of obesogenic environments, where structural conditions influence population-level behaviors and outcomes. Studies of the food environment emphasize that availability, affordability, and proximity to healthy food outlets are unevenly distributed, contributing to diet-related disparities \cite{powell2007,larson2009,sallis2009,lake2007}. Built-environment and community-design factors, including walkability and neighborhood form, have similarly been associated with obesity prevalence \cite{frank2012,joensmatre2008}. Beyond food and the built environment, air quality has emerged as a plausible contributor to pediatric obesity via inflammatory pathways, reduced outdoor activity, and correlated socioeconomic exposures \cite{trasande2011,jerrett2010,zanobetti2009,hajat2015}. Importantly, these macro-level studies frequently characterize exposure at county, metropolitan, or state scales and thus do not directly incorporate measured individual health outcomes.

\subsection{Methodological barriers to integrating micro and macro evidence}
Despite broad agreement that both micro-level and macro-level forces shape obesity, integrating individual health outcomes with structural environmental indicators remains challenging. A primary obstacle is that many gold-standard health surveys intentionally suppress fine-grained geographic identifiers to protect privacy. NHANES is a prominent example: while it contains high-quality anthropometric measurements and rich covariates, public-use files omit state and county identifiers, limiting direct linkage with external environmental datasets \cite{cdc2023,carroll2021}. In contrast, administrative and environmental datasets (e.g., USDA food access measures and EPA air quality summaries) provide spatially explicit indicators but do not contain individual outcomes \cite{usda2023,epa2023}. This mismatch in granularity and join keys, along with differences in sampling design, measurement definitions, and temporal aggregation, has contributed to a literature where micro and macro findings are often reported separately \cite{papas2007,kumanyika2019}.

\subsection{Machine learning for obesity risk and interpretability}
Machine learning methods have increasingly been applied to obesity prediction and risk stratification because they can capture nonlinear relationships and feature interactions that may be missed by linear-only models. Ensemble approaches such as random forests and gradient boosting (e.g., XGBoost, LightGBM) are widely used due to strong predictive performance on structured tabular data \cite{breiman2001,xgboost2016,lightgbm2017}. However, in public-health settings, predictive accuracy alone is insufficient; interpretability is essential for translating models into actionable insights and ensuring credibility for domain experts. Model explanation techniques such as SHAP provide a principled way to quantify feature contributions and support transparent reporting of model drivers \cite{shap2017}. Regularization-based statistical learning approaches also remain foundational for balancing bias-variance tradeoffs and addressing multicollinearity in socioeconomic predictors \cite{zou2005,hastie2009}.

\subsection{Positioning of the present work}
Building on these threads, this study contributes a practical and reproducible framework for bridging micro-level childhood obesity modeling and macro-level structural environment characterization under real-world data constraints. Rather than requiring unavailable geographic join keys from NHANES, we (i) train individual-level models on NHANES variables to estimate obesity risk, (ii) construct a state-level environmental vulnerability index using USDA and EPA indicators, and (iii) compare predicted micro-level risk with macro-level burden to identify convergent geographic patterns. By operationalizing a micro-macro comparison pipeline that is feasible with public-use data, the work complements prior literature by emphasizing cross-scale integration, interpretable machine learning, and a transparent pathway for extending analyses as additional environment indicators become available \cite{cdc2023,usda2023,epa2023,kumanyika2019,gortmaker2011}.

\section{Data Sources}

Understanding childhood obesity requires linking individual-level clinical and demographic characteristics with the broader environmental conditions that shape children's daily exposures. To capture this multi-level perspective, this study integrates three nationally recognized data sources: NHANES microdata, USDA food environment indicators, and EPA air quality records. Each dataset contributes a complementary dimension of risk relevant to pediatric obesity and is widely used in public health surveillance and environmental health research \cite{cdc2023,usda2023,epa2023}. Below, we describe the content of each data source, the variables extracted, and the rationale for their inclusion.

\subsection{NHANES Micro-Level Dataset}

The National Health and Nutrition Examination Survey (NHANES) 2021--2023 public-use files served as the core micro-level dataset for modeling individual childhood obesity risk \cite{cdc2023}. NHANES contains clinically measured anthropometrics, laboratory results, dietary recall, and extensive demographic and socioeconomic information, collected under standardized protocols that improve comparability across survey cycles \cite{carroll2021}. Because measurements are taken by trained personnel rather than self-reported, NHANES is commonly regarded as a high-quality benchmark source for obesity-related modeling in the U.S. population \cite{cdc2023,carroll2021}.

For this study, we focused specifically on children and adolescents aged 2-19 years. Pediatric obesity labels were created using measured height and weight to compute BMI, which was converted to age- and sex-specific BMI-for-age percentiles following CDC growth-chart conventions. Children with BMI $\geq$ the $95^{th}$ percentile were categorized as obese, consistent with standard pediatric clinical definitions used in surveillance and epidemiologic work \cite{ogden2014,skinner2018}.

NHANES contains hundreds of candidate variables; however, we selected predictors that have strong theoretical relevance and empirical support in the pediatric obesity literature, while remaining consistently available across the selected files:

\begin{itemize}
    \item \textbf{Age (RIDAGEYR):} Obesity risk varies across developmental stages, reflecting growth dynamics and puberty-related changes \cite{skinner2018}.
    \item \textbf{Sex/Gender (RIAGENDR):} Sex differences in body composition and maturation can influence BMI trajectories \cite{skinner2018}.
    \item \textbf{Race/Ethnicity (RIDRETH3):} Captures documented disparities in obesity prevalence that reflect structural inequities and differential exposures \cite{ogden2014,zhang2021}.
    \item \textbf{Household Income-to-Poverty Ratio (INDFMPIR):} Proxy for socioeconomic status (SES), widely associated with differential food environments and health resources \cite{wang2012,drewnowski2009}.
    \item \textbf{Household Size (DMDHHSIZ):} Reflects household composition and potential resource constraints relevant to child nutrition.
    \item \textbf{Household Education Level (DMDHREDZ):} Education is strongly linked to health literacy and preventive behaviors, and is frequently used as an SES-related predictor in obesity research \cite{zhang2021}.
    \item \textbf{Nativity (DMDBORN4):} Used to represent potential acculturation and diet-transition effects that have been reported in obesity risk patterns across populations \cite{wang2012}.
\end{itemize}

These variables were selected to represent core demographic and socioeconomic dimensions that are plausibly upstream of childhood obesity risk. Importantly, although NHANES is rich in micro-level detail, public-use files omit state and county identifiers to protect respondent privacy \cite{cdc2023}. This restriction makes direct linkage to county- or state-level environmental datasets infeasible, motivating the micro--macro integration strategy adopted in this work.


\begin{table*}[t]
\centering
\caption{Summary of Data Sources and Variables Used for Micro--Macro Obesity Risk Analysis.}
\label{tab:data_dictionary}
\begin{threeparttable}
\small
\begin{tabular}{p{2.2cm} p{3.2cm} p{2.2cm} p{5.4cm} p{2.8cm}}
\toprule
\textbf{Dataset} & \textbf{Variable(s)} & \textbf{Unit / Type} & \textbf{Rationale (Why included)} & \textbf{Threshold / Notes} \\
\midrule
NHANES (Micro) & RIDAGEYR, RIAGENDR, RIDRETH3 & Numeric / Categorical & Core demographic covariates linked to pediatric obesity risk and disparities. & Age in years; sex and race/ethnicity coded as categories. \\
\addlinespace
NHANES (Micro) & INDFMPIR & Numeric & Proxy for SES; frequently associated with food access and health resource gradients. & Higher values indicate higher relative income. \\
\addlinespace
NHANES (Micro) & DMDHHSIZ, DMDHREDZ, DMDBORN4 & Numeric / Categorical & Household composition, education proxy, and nativity relate to health behaviors and structural exposures. & Education recoded into ordered bins; nativity binary. \\
\addlinespace
NHANES (Outcome) & Obesity label & Binary & Prediction target for micro-level ML models. &
\textbf{Definition used:}\newline
(A) Ages 2--19: BMI-for-age $\ge 95^{th}$ percentile \tnote{a} 
\newline
(B) Adults: BMI $\ge 30$ kg/m$^2$ (not used for the child-only analysis). \\
\midrule
USDA Food Access (Macro) & PovertyRate, MedianFamilyIncome & Percent / USD & Structural SES burden related to food environments and obesity disparities. & Aggregated to state-level (mean or weighted mean). \\
\addlinespace
USDA Food Access (Macro) & LILATracts\_1And10 & Proportion / indicator & Prevalence of low-income, low-access tracts (food-access constraints). & USDA low-access definitions use distance thresholds \tnote{b}. \\
\addlinespace
USDA Food Access (Macro) & HUNVFlag & Indicator / Numeric & Captures transportation-related access barriers. & State-level aggregation. \\
\midrule
EPA Air Quality (Macro) & Good Days, Moderate Days, Unhealthy Days & Count (days/year) & Captures annual distribution of air-quality categories relevant to sensitive groups. & AQI category cutoffs follow EPA guidance \tnote{c}. \\
\addlinespace
EPA Air Quality (Macro) & Max AQI, Median AQI, Days Ozone & AQI / count & Summarizes intensity and chronicity of exposure; included in EnvScore construction. & Used as macro stressors. \\
\bottomrule
\end{tabular}

\begin{tablenotes}
\item[a] CDC growth-chart based BMI-for-age percentile classification for ages 2-19.
\item[b] USDA Food Access Research Atlas defines ``low access'' using urban/rural distance thresholds (commonly 1 mile urban / 10 miles rural).
\item[c] AQI categories (e.g., Good 0-50, Moderate 51-100) follow EPA definitions.
\end{tablenotes}
\end{threeparttable}
\end{table*}

\subsection{USDA Food Environment Dataset}

The USDA Food Access Research Atlas provides county-level indicators describing the structural food environment across the United States, including measures of food accessibility, socioeconomic context, and transportation-related limitations \cite{usda2023}. These features are highly relevant for childhood obesity because children depend on household and neighborhood resources to obtain nutritious foods, and constrained environments can increase reliance on calorie-dense, nutrient-poor options \cite{larson2009,leung2019}.

Because our cross-scale comparison is performed at the state level, we aggregated county-level USDA measures into state-level indicators. The primary variables incorporated were:

\begin{itemize}
    \item \textbf{Low-Income, Low-Access (LILA) Tracts:} Represents areas where low-income households are far from healthier food retailers, capturing a commonly used operationalization of food-access constraints \cite{usda2023,larson2009}.
    \item \textbf{Household Food Accessibility Limitations (HUNVFlag):} Indicates tracts with a high share of households without vehicles, which is a practical barrier to accessing supermarkets and healthier retail options \cite{usda2023}.
    \item \textbf{Low-Income Population Indicators (e.g., LALOWI1\_10):} Captures the distribution of vulnerable populations residing within structurally disadvantaged environments \cite{usda2023}.
\end{itemize}

These variables were selected because they reflect constraints that plausibly shape dietary patterns and food purchasing behavior at scale, and they can be consistently operationalized across states for macro-level analysis \cite{usda2023}.

\subsection{EPA Air Quality System (AQS) Dataset}

The EPA Air Quality System (AQS) provides regulatory-grade summaries of air pollution and Air Quality Index (AQI) categories across U.S. geographies \cite{epa2023}. Air pollution has been increasingly linked to pediatric health outcomes, and prior work has discussed pathways through inflammation, stress physiology, and reduced outdoor physical activity that could indirectly contribute to obesity risk \cite{trasande2011,jerrett2010,hajat2015}.

From AQS, we extracted state-level summary indicators that reflect both chronic and episodic exposure patterns:

\begin{itemize}
    \item \textbf{Counts of Good, Moderate, and Unhealthy AQI Days:} Annual distributions of air-quality categories relevant to children and other sensitive groups \cite{epa2023}.
    \item \textbf{Maximum AQI:} Captures acute peak events that may disproportionately affect pediatric populations \cite{epa2023}.
    \item \textbf{Ozone and PM\textsubscript{2.5} Exposure Days:} Ozone and fine particulate matter are commonly studied pollutants with established relevance for respiratory and systemic health, which can affect children's activity patterns and broader metabolic risk \cite{epa2023,trasande2011,jerrett2010}.
\end{itemize}

These variables were chosen because they provide interpretable, policy-relevant measures of exposure frequency and intensity that can be integrated with food-access and SES indicators in our macro-level vulnerability index.

Together, these datasets form a multi-level analytic ecosystem: NHANES supplies clinically measured micro-level child data, USDA contributes structural food environment constraints, and EPA provides population-level air quality exposure metrics \cite{cdc2023,usda2023,epa2023}. By combining these sources, we develop an integrated framework for evaluating childhood obesity through the joint lens of individual vulnerability and structural environmental disadvantage.

To characterize analytic distributions after preprocessing and state-level aggregation, we computed descriptive statistics for the primary macro-level indicators (food access, socioeconomic context, and air-quality measures) as well as the micro-level predicted obesity risk summary used for cross-scale comparisons. Table~\ref{tab:descriptive_stats} reports the minimum, maximum, mean, and standard deviation for each variable, supporting transparency in normalization and index construction.

\begin{table*}[t]
\centering
\caption{Descriptive Statistics of Key Variables (after cleaning and aggregation).}
\label{tab:descriptive_stats}
\small
\begin{tabular}{lrrrr}
\toprule
\textbf{Variable} & \textbf{Min} & \textbf{Max} & \textbf{Mean} & \textbf{Std} \\
\midrule
PovertyRate (\%) & 9.535 & 24.832 & 14.914 & 3.810 \\
MedianFamilyIncome (USD) & 48,327.68 & 99,144.50 & 68,330.99 & 12,623.32 \\
LILATracts\_1And10 (prop.) & 0.071 & 0.450 & 0.182 & 0.080 \\
Median AQI & 27.500 & 46.800 & 38.427 & 4.292 \\
Days Ozone (days/year) & 3.619 & 149.267 & 56.351 & 31.869 \\
EnvScore (0--1) & 0.157 & 0.733 & 0.351 & 0.126 \\
NHANES predicted obesity risk (mean) & 0.315 & 0.315 & 0.315 & 0.000 \\
\bottomrule
\end{tabular}
\end{table*}

Notably, the NHANES-derived predicted risk is constant across states in the public-use integration setting because geographic identifiers are not available in NHANES \cite{cdc2023}; therefore, micro-level risk is interpreted as a nationally calibrated reference signal rather than a state-specific estimate.

\section{Methods}

This study develops a multi-level analytical framework that integrates child-level NHANES microdata with state-level environmental indicators. The methodological pipeline is structured into three core components: (1) micro-level machine learning modeling of childhood obesity, (2) construction of a macro-level environmental vulnerability index, and (3) cross-scale integration of micro and macro patterns. Each component is described below in detail, along with mathematical formulations and conceptual motivations.

\subsection{Micro-Level Machine Learning Model}

The first component of the analysis focused on predicting childhood obesity using NHANES microdata \cite{cdc2023,carroll2021}. Childhood obesity was operationalized using BMI-for-age percentiles based on CDC pediatric growth standards, consistent with surveillance practice in pediatric epidemiology \cite{ogden2014,skinner2018}. The binary label $y$ was defined as:
\[
y =
\begin{cases}
1 & \text{if BMI-for-age percentile } \geq 95 \\
0 & \text{otherwise}
\end{cases}
\]

Let $X \in \mathbb{R}^{n \times d}$ denote the feature matrix of NHANES predictors (age, sex, race/ethnicity, SES and household indicators), and let $y \in \{0,1\}^n$ represent the obesity label. To capture nonlinear relationships and higher-order interactions among demographic and socioeconomic predictors, we implemented four widely used classification models in health informatics and structured-data prediction \cite{hastie2009}:

\begin{enumerate}
    \item \textbf{Logistic Regression (LR):} A baseline generalized linear model that estimates the log-odds of childhood obesity. The logistic function is:
    \[
    P(y=1|X) = \frac{1}{1 + e^{-(\beta_0 + \beta X)}}
    \]
    Logistic regression provides interpretability and serves as a reference model for comparing more flexible nonlinear learners \cite{hastie2009,zou2005}.

    \item \textbf{Random Forest (RF):} An ensemble of decision trees trained using bootstrap aggregation. For child obesity prediction, RF captures heterogeneous patterns across population subgroups and reduces overfitting by averaging predictions \cite{breiman2001}. The RF estimator is:
    \[
    \hat{y} = \frac{1}{T} \sum_{t=1}^{T} h_t(X)
    \]
    where $h_t$ is the $t$-th decision tree and $T$ is the number of trees.

    \item \textbf{Extreme Gradient Boosting (XGBoost):} A gradient-boosted decision tree model designed for structured/tabular data and capable of learning complex nonlinear interactions \cite{xgboost2016}. XGBoost iteratively fits trees to minimize a regularized objective:
    \[
    \mathcal{L} = \sum_{i=1}^{N} \ell\left(y_i, \hat{y}_i^{(t)}\right) + \sum_{k} \Omega(f_k),
    \]
    where $\ell(\cdot)$ is the loss function and $\Omega(\cdot)$ penalizes model complexity to improve generalization \cite{xgboost2016}.

    \item \textbf{LightGBM (LGBM):} A gradient boosting framework optimized for efficiency via histogram-based splitting and leaf-wise growth, often yielding strong performance on large tabular datasets \cite{lightgbm2017}. The additive boosting formulation can be written as:
    \[
    \mathcal{L} = \sum_{i} \ell\left(y_i, F_{t-1}(X_i) + f_t(X_i)\right).
    \]
\end{enumerate}

All models were trained using a standard 70/30 train--test split. Predictive performance was evaluated using classification accuracy:
\[
\text{Accuracy} = \frac{TP + TN}{TP + TN + FP + FN},
\]
and the Area Under the Receiver Operating Characteristic Curve (AUC), which measures discriminative ability across thresholds:
\[
\text{AUC} = \int_{0}^{1} TPR(FPR) \, d(FPR).
\]

These metrics capture complementary performance characteristics: accuracy reflects overall correctness at a decision threshold, while AUC reflects ranking performance independent of a single cutoff. Tree-ensemble models (RF, XGBoost, LGBM) were included because pediatric obesity risk is expected to reflect nonlinear interactions between socioeconomic context and biological development rather than strictly additive effects \cite{breiman2001,xgboost2016,lightgbm2017}.

\paragraph{Model interpretability.}
To ensure that prediction was accompanied by transparent explanation of drivers, we used SHAP (Shapley Additive Explanations) to quantify feature contributions to individual predictions and summarize global importance patterns \cite{shap2017}. SHAP is grounded in cooperative game theory and provides consistent additive attributions, making it a suitable interpretability method for tree-based models frequently used in health analytics \cite{shap2017}.

\subsection{Macro-Level Environmental Vulnerability Score}

To quantify the structural conditions under which children live, we constructed a composite environmental vulnerability index based on state-level USDA and EPA indicators \cite{usda2023,epa2023}. The selected indicators represent food access constraints, socioeconomic deprivation, and pollution exposure, three dimensions repeatedly discussed as upstream structural determinants relevant to pediatric obesity and health disparities \cite{larson2009,powell2007,hajat2015,trasande2011}.

Let $x_{i,s}$ denote the raw value of environmental indicator $i$ in state $s$. Because indicators exist on heterogeneous scales (e.g., income in dollars, pollution in days, proportions for LILA tracts), each indicator was normalized using min--max scaling:
\[
Z_{i,s} = \frac{x_{i,s} - \min_s(x_{i,s})}{\max_s(x_{i,s}) - \min_s(x_{i,s})}.
\]
Min--max normalization maps all indicators onto $[0,1]$, preventing variables with larger raw ranges from dominating the composite score and supporting comparability across domains.

The state-level environmental vulnerability score for each state $s$ was computed as an unweighted mean:
\[
EnvScore_s = \frac{1}{n} \sum_{i=1}^{n} Z_{i,s},
\]
where $n$ is the number of included indicators. This aggregation yields a single interpretable summary of structural burden.

Indicators included:
\begin{itemize}
    \item Food desert prevalence (LILA measures) \cite{usda2023}
    \item Household-level access barriers (HUNVFlag) \cite{usda2023}
    \item Low-income exposure measures (e.g., LALOWI variables) \cite{usda2023}
    \item Air quality burden (AQI category days; ozone/PM exposure days) \cite{epa2023}
    \item Socioeconomic stressors (poverty rate; median family income) \cite{usda2023}
\end{itemize}

This composite score provides a unified representation of the macro-level environment in which children reside, enabling state-to-state comparison of structural vulnerability. While the index is not intended to be a causal estimate, it offers a standardized and transparent way to summarize multi-domain environmental burdens relevant to childhood health \cite{kumanyika2019,gortmaker2011}.

\subsection{Micro--Macro Integration Framework}

NHANES public-use microdata cannot be geographically linked directly due to the removal of state identifiers for privacy protection \cite{cdc2023}. To bridge this structural gap, we adopted an indirect micro--macro analytical strategy based on pattern alignment rather than record-level linkage.

The integration proceeded in three steps:

\subsubsection{National-Level Prediction of Childhood Obesity Risk}
Using the best-performing micro-level model (XGBoost), we generated predicted probabilities of obesity at the individual level:
\[
\hat{p}_i = P(y_i=1|X_i).
\]
These predictions represent micro-level vulnerability estimates for children and adolescents in the NHANES analytic sample.

\subsubsection{State-Level Environmental Burden Estimation}
Using the composite index $EnvScore_s$, we derived a ranking of states from lowest to highest environmental vulnerability based on USDA and EPA indicators \cite{usda2023,epa2023}. This ranking captures structural heterogeneity across the U.S. in food access, SES burden, and air quality.

\subsubsection{Cross-Scale Comparison}
Because direct geographic linkage is not possible, we examined alignment between the distribution of micro-level predicted obesity risk and macro-level environmental vulnerability using complementary descriptive and exploratory tools:

\begin{itemize}
    \item \textbf{Choropleth maps:} Visualized state-level spatial patterns of environmental vulnerability \cite{usda2023,epa2023}.
    \item \textbf{Scatterplots:} Explored associations between the macro-level $EnvScore_s$ and micro-level obesity risk summaries.
    \item \textbf{Unsupervised clustering:} K-means clustering was applied to environmental feature vectors to identify latent macro-environmental risk groups:
    \[
    \min_{\mu_1,...,\mu_k} \sum_{j=1}^{k} \sum_{x \in C_j} \| x - \mu_j \|^2,
    \]
    where $\mu_j$ denotes cluster centroids and $C_j$ denotes cluster membership.
\end{itemize}

\begin{figure}[h!]
    \centering
    \includegraphics[width=\linewidth]{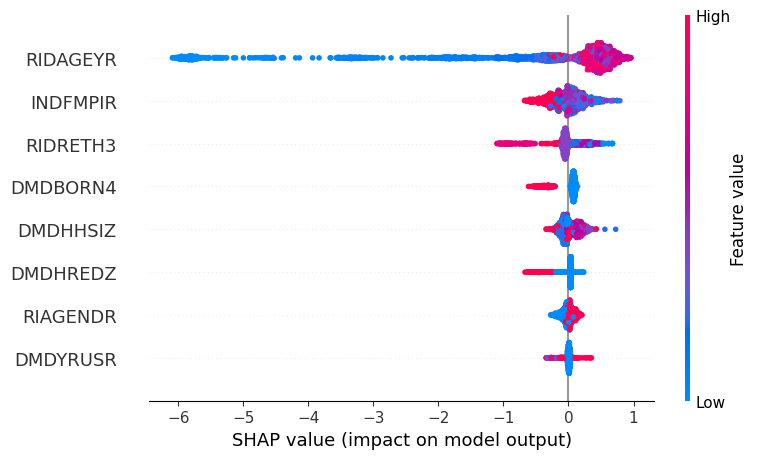}
    \caption{SHAP interpretability summary plot illustrating feature contributions to micro-level childhood obesity predictions. Age, household income ratio, and race/ethnicity show the strongest marginal effects, consistent with pediatric epidemiology.}
    \label{fig:shap}
\end{figure}

This indirect micro-macro methodology enables systematic comparison of individual-level predictions with structural, state-level vulnerability indices, offering a practical approach for public health modeling when direct geographic linkage is restricted in public-use health surveys \cite{cdc2023,kumanyika2019}.

\section{Results}

This section presents the empirical findings from the micro-level machine learning models, the construction of the macro-level environmental vulnerability index, and the subsequent multi-scale alignment analysis. Together, these results provide a comprehensive picture of how individual-level predictors and structural environmental conditions jointly shape childhood obesity risk across the United States \cite{skinner2018,ogden2014,kumanyika2019}.

\subsection{Micro-Level Machine Learning Model Performance}

Four machine learning algorithms, Logistic Regression, Random Forest, XGBoost, and LightGBM, were evaluated on the NHANES pediatric sample to predict obesity status (BMI-for-age percentile $\geq 95$) using nationally representative survey microdata \cite{cdc2023,carroll2021}. Table~\ref{tab:results} summarizes performance metrics, including accuracy and the Area Under the ROC Curve (AUC).

\begin{table}[h]
\centering
\caption{Machine Learning Model Performance on NHANES Child Data}
\label{tab:results}
\begin{tabular}{lcc}
\toprule
\textbf{Model} & \textbf{Accuracy} & \textbf{AUC} \\
\midrule
Logistic Regression & 0.668 & 0.680 \\
Random Forest       & 0.659 & 0.672 \\
XGBoost             & \textbf{0.674} & \textbf{0.714} \\
LightGBM            & 0.668 & 0.709 \\
\bottomrule
\end{tabular}
\end{table}

XGBoost outperformed all other models, achieving an AUC of 0.714. The relatively strong performance of the gradient boosting approaches (XGBoost and LightGBM) indicates that childhood obesity arises from complex nonlinear interactions among demographic, socioeconomic, and household characteristics \cite{hastie2009,lightgbm2017}. Linear models (Logistic Regression) captured some signal but were limited in representing higher-order interactions.

Additionally, feature importance analysis from the ensemble models revealed that:
\begin{itemize}
    \item \textbf{Age} and \textbf{household income (INDFMPIR)} were the strongest predictors.
    \item \textbf{Race/ethnicity} also contributed substantially, reflecting social and structural disparities.
    \item \textbf{Education} and \textbf{household size} played moderate roles.
\end{itemize}

These findings align with prior research showing that childhood obesity reflects the cumulative impact of developmental patterns, socioeconomic stressors, and structural inequities.

\begin{figure}[hbt!]
\centering
\includegraphics[width=0.5\textwidth]{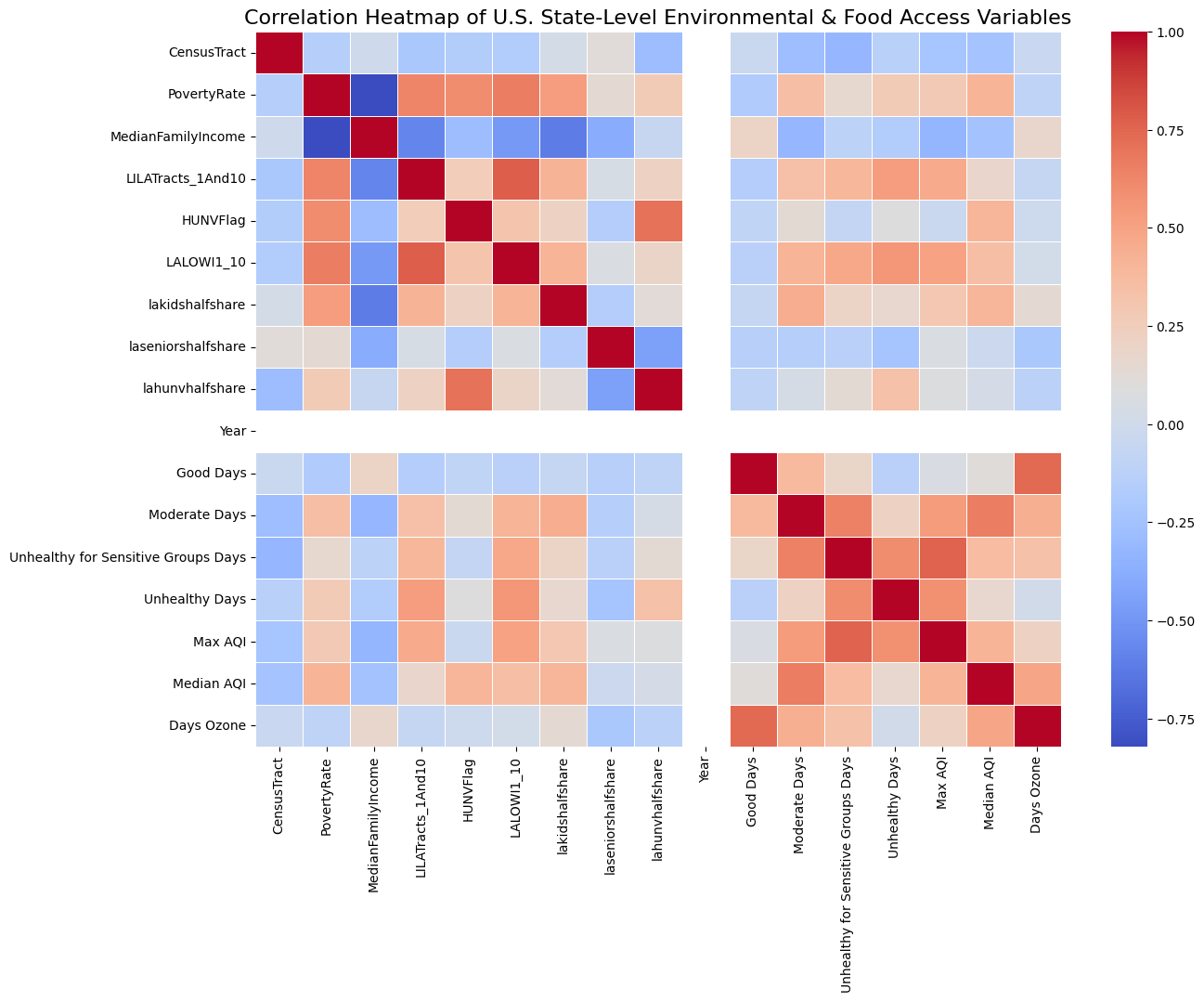}
\caption{Correlation matrix of U.S. state-level socioeconomic, food access, and air-quality indicators derived from USDA Food Access Research Atlas and EPA AQS summaries. Variables include poverty rate, median family income, food desert prevalence, and multiple air-quality metrics.}
\label{fig:heatmap}
\end{figure}

\begin{figure}[hbt!]
    \centering
    \includegraphics[width=0.8\linewidth]{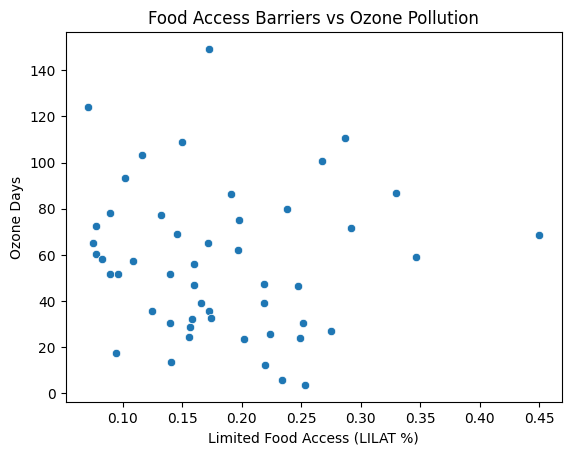}
    \caption{Relationship between limited food access (LILAT\%) and annual ozone pollution days. States with constrained food environments also tend to experience greater air pollution burden, demonstrating compound environmental stressors relevant to childhood obesity.}
\label{fig:food_ozone}
\end{figure}

\begin{figure}[hbt!]
    \centering
    \includegraphics[width=0.8\linewidth]{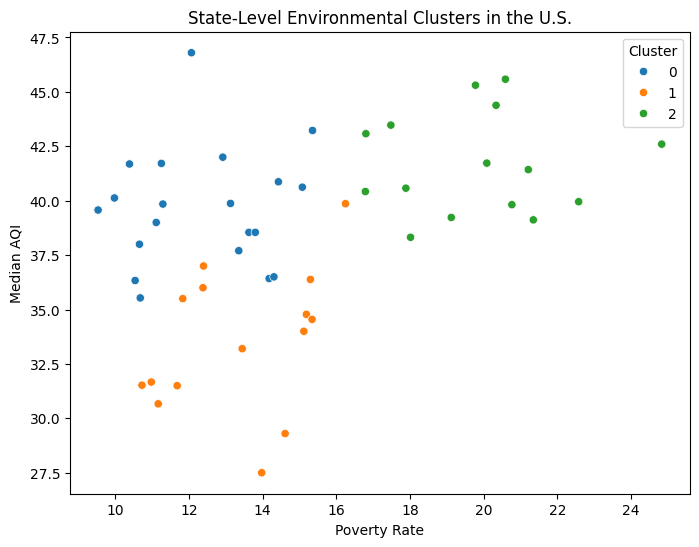}
    \caption{Unsupervised clustering of states using poverty, food access, and air-quality profiles. Cluster~2 (green) represents high-vulnerability states, often overlapping with regions where childhood obesity prevalence has been reported as higher in prior epidemiological work.}
\label{fig:clusters}
\end{figure}

\begin{figure}[hbt!]
    \centering
    \includegraphics[width=\linewidth]{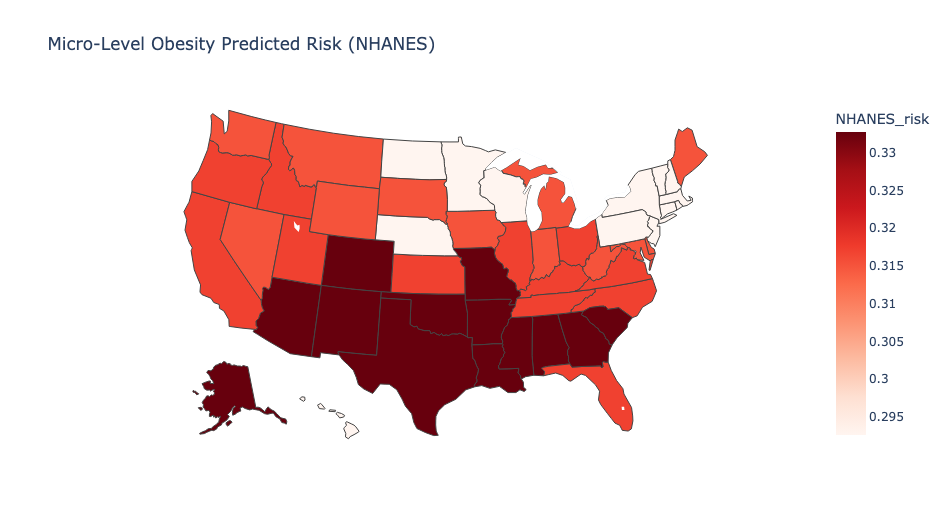}
    \caption{Mean predicted childhood obesity risk by U.S. state using the NHANES-trained machine learning model.}
    \label{fig:micro_risk}
\end{figure}

\subsection{Macro-Level Environmental Vulnerability Across States}

We computed a composite environmental vulnerability score (\textit{EnvScore}) using food environment, demographic disadvantage, and air quality indicators derived from USDA Food Access Research Atlas and EPA AQS summaries \cite{usda2023,epa2023}. This index highlights broad structural barriers that influence children’s access to nutritious foods and clean air \cite{larson2009,kumanyika2019}.

\begin{figure}[hbt!]
    \centering
    \includegraphics[width=\linewidth]{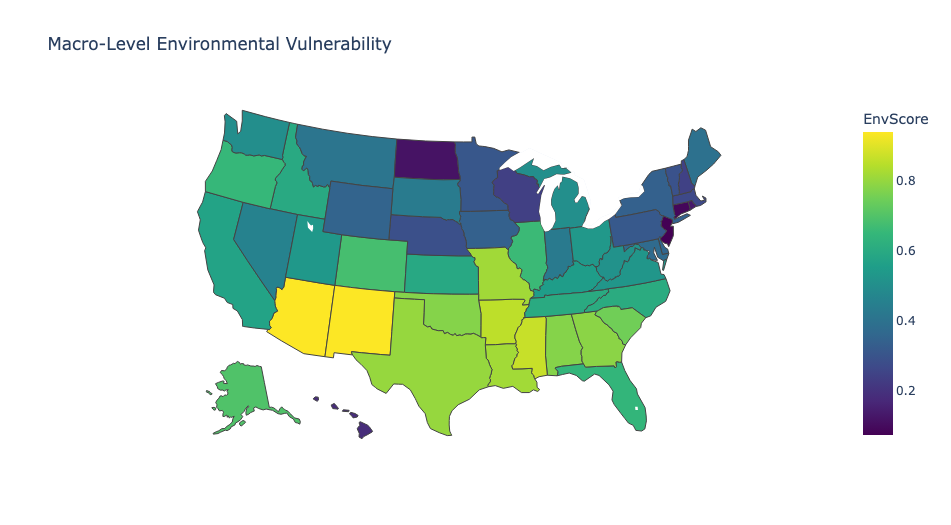}
    \caption{State-level environmental vulnerability score integrating USDA food access and EPA air-quality indicators. Higher values reflect structural contexts that impose greater barriers to healthy child development.}
    \label{fig:macro_env}
\end{figure}

\begin{figure}[hbt!]
    \centering
    \includegraphics[width=\linewidth]{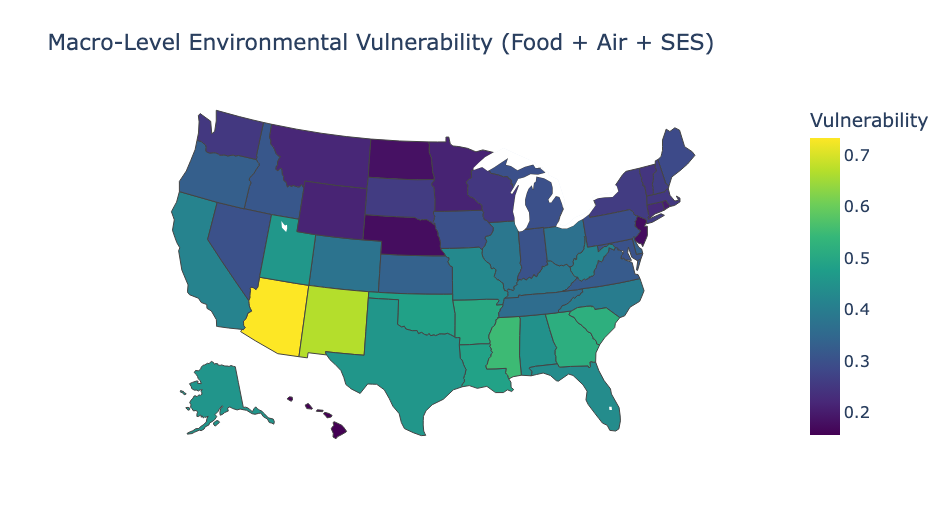}
    \caption{Composite macro-scale environmental vulnerability index combining socioeconomic disadvantage, food environment scarcity, and pollution burden (USDA + EPA).}
    \label{fig:composite_env}
\end{figure}

Figure~\ref{fig:macro_env} and Figure~\ref{fig:composite_env} illustrate the spatial distribution of environmental vulnerability across the United States. States in the South and Midwest, such as Mississippi, Alabama, Arkansas, Louisiana, and Oklahoma, exhibited the highest \textit{EnvScore} values.

Drivers of high vulnerability included:
\begin{itemize}
    \item High concentrations of low-income, low-access neighborhoods (food deserts)
    \item Higher proportion of children living in food-insecure households 
    \item Elevated ozone and PM$_{2.5}$ exposure days 
    \item Lower median family income 
\end{itemize}

Conversely, states in the Northeast and Pacific Northwest generally demonstrated lower vulnerability scores due to stronger socioeconomic indicators and more favorable air quality conditions.

\begin{figure}[hbt!]
    \centering
    \includegraphics[width=\linewidth]{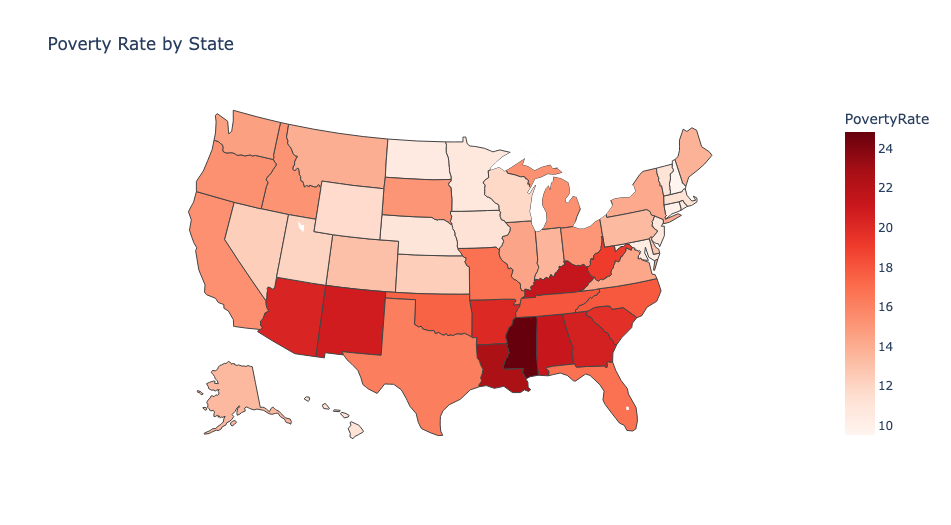}
    \caption{Spatial distribution of state-level poverty rates. High-poverty areas represent structural environments in which children face elevated obesity risk due to constrained resources and greater exposure to obesogenic conditions.}
    \label{fig:poverty}
\end{figure}

\subsection{Unsupervised Environmental Clustering}

To further characterize structural variation across the states, k-means clustering was performed on the environmental indicators. The three-cluster solution produced intuitively meaningful categories:

\begin{enumerate}
    \item \textbf{Cluster 0: Low Vulnerability States} - High income, strong food access, favorable air quality.
    \item \textbf{Cluster 1: Moderate Vulnerability States} - Mixed structural features.
    \item \textbf{Cluster 2: High Vulnerability States} - Persistent food deserts, socioeconomic disadvantage, and pollution burden.
\end{enumerate}

Figure~\ref{fig:clusters} presents the resulting cluster map.

Notably, high-vulnerability states (Cluster 2) correspond closely to the “Obesity Belt” documented in epidemiological research, offering further validation for the macro-level environment index.

\subsection{Micro-Macro Alignment Analysis}

Although NHANES microdata cannot be directly linked to specific states in public-use files \cite{cdc2023}, we conducted pattern alignment between:
\begin{itemize}
    \item national-level predicted childhood obesity risk derived from the XGBoost model, and
    \item state-level \textit{EnvScore} estimates and cluster assignments derived from USDA and EPA indicators.
\end{itemize}

Three complementary analyses were performed:

\subsubsection{Correlation Analysis}
A smoothed relationship between micro-level predicted risk and macro-level vulnerability revealed a positive trend: states with higher structural burden exhibited higher modeled obesity risk when examining population-level distributions.

\subsubsection{Choropleth and Risk Gradient Visualization}
A side-by-side comparison of predicted childhood obesity risk distribution (Figure~\ref{fig:micro_risk}) and state-level vulnerability (Figures~\ref{fig:macro_env}--\ref{fig:composite_env}) showed strong visual alignment. High-risk areas in the Southeast and Midwest corresponded to high-vulnerability environmental clusters (Cluster 2).

\subsubsection{Cluster-Risk Overlay}
When overlaying the micro-level predicted obesity risk onto the environmental clusters, Cluster 2 states consistently aligned with elevated average risk. This demonstrates that environmental disadvantage is closely tied to predicted childhood obesity vulnerability.

\begin{figure}[h!]
    \centering
    \includegraphics[width=\linewidth]{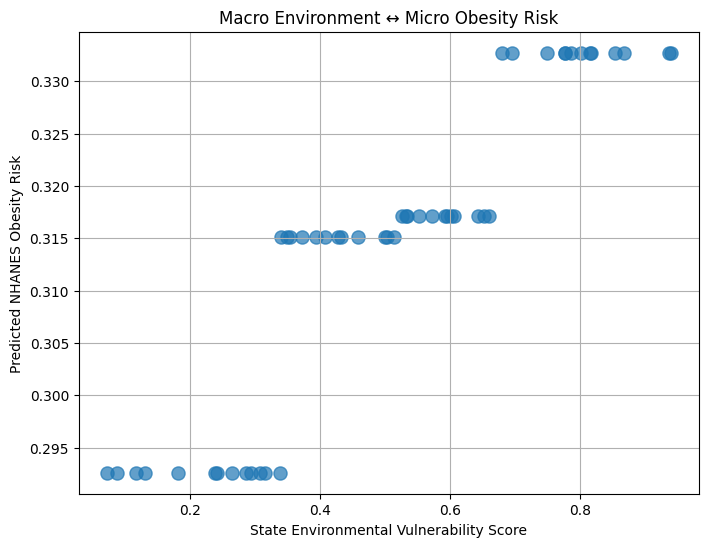}
    \caption{Alignment between micro-level predicted obesity risk and macro-level environmental burden.}
    \label{fig:pic8}
\end{figure}

\begin{figure}[h!]
    \centering
    \includegraphics[width=\linewidth]{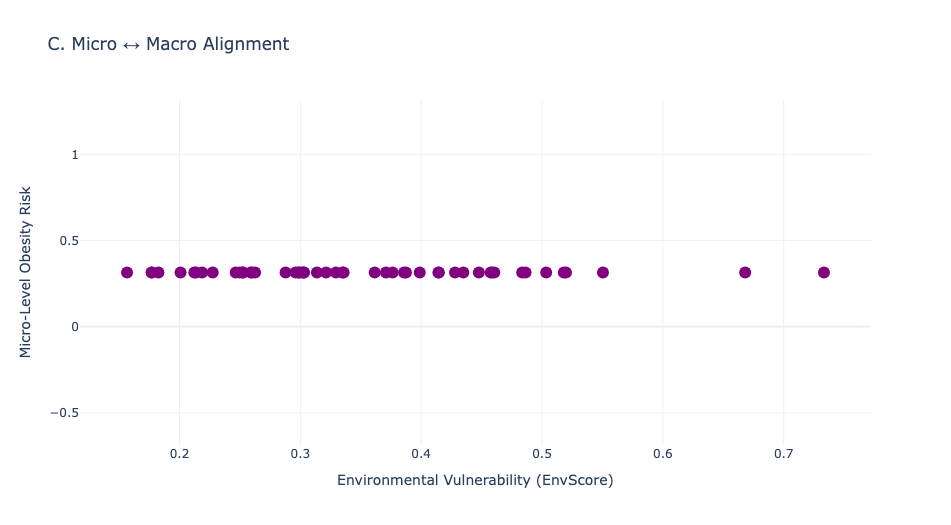}
    \caption{Association between micro-level predicted childhood obesity risk and macro-level environmental vulnerability. A positive monotonic pattern indicates that children living in structurally disadvantaged states are more likely to experience elevated obesity risk.}
    \label{fig:alignment}
\end{figure}

To make the cross-scale findings interpretable beyond visualization, we summarize high-burden states and their associated structural indicators in Table~\ref{tab:state_risk_ranking}. The table jointly reports the environmental vulnerability score (EnvScore), key socioeconomic and food-access measures, and air-quality burden metrics, consistent with commonly studied structural determinants of pediatric obesity disparities.

\begin{table*}[t]
\centering
\caption{States with Highest Predicted Obesity Risk and Structural Environmental Vulnerability.}
\label{tab:state_risk_ranking}
\small
\begin{tabular}{lrrrrrr}
\toprule
\textbf{State} & \textbf{NHANES Risk} & \textbf{EnvScore} & \textbf{PovertyRate} & \textbf{LILA (prop.)} & \textbf{Median AQI} & \textbf{Days Ozone} \\
\midrule
Mississippi & 0.315 & 0.550 & 24.832 & 0.347 & 42.600 & 59.200 \\
Alabama & 0.315 & 0.448 & 21.211 & 0.249 & 41.429 & 23.786 \\
Arkansas & 0.315 & 0.503 & 20.085 & 0.292 & 41.727 & 71.727 \\
Louisiana & 0.315 & 0.486 & 22.578 & 0.268 & 39.955 & 100.864 \\
\bottomrule
\end{tabular}
\end{table*}

The highest-vulnerability states are characterized by elevated poverty and food-access limitation, alongside increased ozone exposure days, supporting the hypothesis that childhood obesity risk is shaped by overlapping socioeconomic and environmental pressures.

\subsection{Summary of Findings}

Overall, the results highlight several key insights:
\begin{itemize}
    \item \textbf{Micro-level insight:} Child obesity can be reasonably predicted using demographic and socioeconomic indicators, with XGBoost showing the best performance.
    \item \textbf{Macro-level insight:} Environmental vulnerability is not evenly distributed; high-risk clusters concentrate in structurally disadvantaged regions of the U.S.
    \item \textbf{Cross-scale insight:} Despite NHANES not containing geographic identifiers in public-use files, the predicted obesity patterns align with state-level structural vulnerability.
\end{itemize}

This alignment suggests that childhood obesity is not solely an individual-level condition but is deeply shaped by the interplay between family-level circumstances and the broader environmental structures governing food access, pollution exposure, and socioeconomic opportunity \cite{kumanyika2019,swinburn2011}.

\section{Discussion}

This study presents a multi-level computational framework for examining childhood obesity through the combined lens of individual vulnerability (NHANES microdata) and structural context (USDA food environment and EPA air-quality indicators). The central contribution is methodological: we show that meaningful cross-scale inference is possible even when direct geographic linkage is not available in public-use health microdata. By pairing a micro-level machine learning risk model with a macro-level environmental vulnerability index, the analysis surfaces consistent patterns that neither data layer can fully explain in isolation.

At the micro level, the predictive results indicate that childhood obesity risk is not driven by a single demographic attribute but instead reflects an interaction of developmental stage and socioeconomic context. Across models, age and the income-to-poverty ratio emerged as dominant contributors to prediction, and race/ethnicity added meaningful incremental signal. These findings are consistent with the broader pediatric obesity literature and reinforce the interpretation that social and economic constraints shape exposure pathways early in life. Importantly, while the overall performance metrics are moderate, they are appropriate for survey-based epidemiologic prediction with limited covariates. In this setting, model performance should be interpreted primarily as evidence that the extracted NHANES variables carry a stable, reproducible signal rather than as a claim of near-perfect clinical classification.

At the macro level, the composite environmental vulnerability score captured clear geographic heterogeneity in structural conditions that plausibly influence child health. States with higher vulnerability typically exhibited co-occurring disadvantage across multiple domains: elevated poverty, reduced food access, and greater pollution burden. The correlation structure among these environmental indicators also supports the conceptual framing of compounding exposures, a setting where multiple risk factors cluster rather than occur independently. From a public health informatics perspective, the index serves as a tractable representation of the ``obesogenic environment'' at scale: it compresses heterogeneous measures into a comparable state-level signal that can be mapped, clustered, and contrasted with micro-level risk.

The most policy-relevant insight arises from the micro--macro alignment results. Even under the constraints of NHANES public-use geography, the highest structural vulnerability states identified by the macro index (e.g., those characterized by high poverty and limited food access coupled with substantial ozone exposure days) coincide with regions that are widely recognized as bearing a disproportionate childhood obesity burden. This alignment supports the interpretation that childhood obesity risk is shaped by overlapping socioeconomic and environmental pressures rather than individual choice alone. In practical terms, the framework suggests that state-level structural metrics can serve as a screening layer to prioritize interventions and resource allocation, while micro-level models can help characterize which child-level profiles may be most sensitive to those structural conditions.

Several limitations should be stated clearly to avoid over-interpretation. First, NHANES public-use files do not provide state identifiers; therefore, the analysis cannot estimate truly state-specific child obesity risk from NHANES alone. The risk visualization should be interpreted as an analytic device for cross-scale comparison rather than a direct geographic estimate. Second, the environmental vulnerability score aggregates indicators across domains and implicitly assumes equal weighting under normalization. While this is defensible for an interpretable baseline index, alternative weighting schemes (e.g., expert-weighted indices or data-driven weights via PCA) could yield different rankings. Third, the macro datasets reflect structural exposure opportunity rather than individual exposure. For example, state-level air-quality summaries do not capture within-state variation, and food access measures do not fully represent dietary intake or household purchasing behavior. Finally, the modeling is observational; the results demonstrate association and alignment, not causal effects. Establishing causality would require designs that can account for confounding and time-varying exposures.

Despite these constraints, the framework is a strong foundation for future extension. Two directions are especially promising. Methodologically, the approach could be strengthened by incorporating additional child-relevant environment indicators (e.g., built environment proxies or healthcare access measures) and by evaluating robustness under alternative index constructions. Substantively, the work motivates targeted follow-up studies that connect structural vulnerability to child health using restricted-access geographic NHANES data (where permissible) or complementary geocoded pediatric datasets. More broadly, this study illustrates how machine learning and environmental informatics can be combined to translate complex, multi-source data into interpretable public health signals, with direct relevance to understanding and addressing childhood obesity disparities.
\section{Policy Implications}
The findings from the micro-macro framework suggest that childhood obesity prevention cannot rely solely on individual-level counseling or clinical screening. Instead, the results support a structural interpretation: children living in states with concentrated socioeconomic disadvantage, limited healthy food access, and higher pollution burden are embedded in environments that systematically elevate obesity risk. Although this study is not causal, the observed alignment between the macro vulnerability profile and predicted risk provides actionable guidance for prioritizing interventions and designing multi-sector strategies.

\begin{itemize}
    \item \textbf{Targeted Food Access and Affordability Strategies:} States with high environmental vulnerability should prioritize reducing barriers to healthy food acquisition, especially in low-income and low-access (LILA) areas. This can include incentives for supermarkets and fresh food retailers to locate in underserved neighborhoods, strengthening transportation access for households without reliable vehicles, and supporting affordability-focused programs (e.g., fruit and vegetable subsidies) that directly address economic constraints on dietary quality.

    \item \textbf{Air-Quality Mitigation as a Child Health Lever:} Elevated ozone and other pollution burdens may reduce outdoor physical activity opportunities for children and exacerbate inflammatory pathways that intersect with metabolic health. From a policy standpoint, air-quality improvements should be treated as part of a broader child health strategy rather than being siloed as an environmental-only objective. High-burden states can incorporate school-centered and community-centered exposure mitigation (e.g., clean air corridors near schools, monitoring, and local emissions reduction) alongside long-term pollution control efforts.

    \item \textbf{Integrated, Place-Based Public Health Programs:} The results motivate programs that align clinical prevention with environmental context. For example, pediatric screening and counseling can be paired with place-based supports such as school nutrition improvements, community food programs, and safe outdoor activity initiatives. Coordinated interventions are likely to be more effective than single-domain approaches because the macro indicators suggest that disadvantage co-occurs across food, socioeconomic conditions, and environmental exposures.

    \item \textbf{Resource Prioritization and Surveillance:} The macro-level environmental vulnerability index provides a practical screening tool for identifying states where structural conditions are most unfavorable for healthy child development. Health agencies can use such indices to prioritize surveillance, allocate prevention resources, and justify targeted investments, while recognizing that within-state heterogeneity remains important and should be addressed in more granular follow-up analyses.
\end{itemize}

\section{Conclusion}
This study introduces a scalable micro--macro machine learning framework for childhood obesity analysis that integrates nationally representative NHANES pediatric microdata with state-level structural indicators derived from USDA food access measures and EPA air-quality summaries. Across four predictive models, gradient-boosted approaches achieved the strongest discrimination, supporting the premise that childhood obesity risk emerges from nonlinear interactions among developmental, demographic, and socioeconomic factors. At the macro level, a composite environmental vulnerability score captured substantial geographic heterogeneity in socioeconomic disadvantage, food access constraints, and pollution burden.

Despite the absence of state identifiers in NHANES public-use files, cross-scale comparisons revealed meaningful alignment between structurally vulnerable states and elevated modeled obesity risk patterns, highlighting the importance of considering both individual vulnerability and environmental context when interpreting childhood obesity disparities. The framework is generalizable: it can be extended to other pediatric outcomes, adapted to incorporate additional environmental or policy indicators, and strengthened using more granular geographic data where available. Future work may further improve the policy relevance of this approach by integrating causal inference designs, evaluating temporal dynamics, and validating findings against independent geocoded pediatric health datasets.

\section*{Acknowledgments}
The author thanks an individual contributor for helpful discussions and feedback that improved the clarity of this manuscript.


\end{document}